\documentclass[sigconf]{acmart}
\usepackage{ctable}
\AtBeginDocument{%
  \providecommand\BibTeX{{%
    \normalfont B\kern-0.5em{\scshape i\kern-0.25em b}\kern-0.8em\TeX}}}

\usepackage{amsmath}
\usepackage{bbm}
\usepackage{enumitem}
\usepackage{mathtools,xparse}
\usepackage{subfigure}

\DeclarePairedDelimiter{\norm}{\lVert}{\rVert}

\usepackage{kotex}

\newcommand{\Poincare}{Poincar\'e}
\renewcommand{\shortauthors}{Kyuyong et al.}
\graphicspath{{fig/}}
\setcopyright{acmcopyright}
\copyrightyear{2020}
\acmYear{2020}
\acmDOI{10.1145/1122445.1122456}

\acmConference[AdKDD'20]{AdKDD'20}{August 23, 2020}{San Diego, CA, USA}
\acmBooktitle{AdKDD'20, August 23, 2020, San Diego, CA, USA}
\acmPrice{15.00}
\acmISBN{978-1-4503-XXXX-X/18/06}

\begin{document}

\title{Multi-Manifold Learning for Large-Scale Targeted Advertising System}

\author{Kyuyong Shin}
\email{ky.shin@navercorp.com}
\affiliation{%
  \institution{Clova AI Research, NAVER Corp.}
}

\author{Young-Jin Park}
\email{young.j.park@navercorp.com}
\affiliation{%
  \institution{Naver R\&D Center, NAVER Corp.}
}

\author{Kyung-Min Kim}
\email{kyungmin.kim.ml@navercorp.com}
\affiliation{%
  \institution{Clova AI Research, NAVER Corp.}
}

\author{Sunyoung Kwon}
\email{sunny.kwon@navercorp.com}
\affiliation{%
  \institution{Clova AI Research, NAVER Corp.}
}

\begin{abstract}
Messenger advertisements (ads) give direct and personal user experience yielding high conversion rates and sales. However, people are skeptical about ads and sometimes perceive them as spam, which eventually leads to a decrease in user satisfaction.
Targeted advertising, which serves ads to individuals who may exhibit interest in a particular advertising message, is strongly required. The key to the success of precise user targeting lies in learning the accurate user and ad representation in the embedding space.
Most of the previous studies have limited the representation learning in the Euclidean space, but recent studies have suggested hyperbolic manifold learning for the distinct projection of complex network properties emerging from real-world datasets such as social networks, recommender systems, and advertising.
We propose a framework that can effectively learn the hierarchical structure in users and ads on the hyperbolic space, and extend to the \emph{Multi-Manifold Learning}. Our method constructs multiple hyperbolic manifolds with learnable curvatures and maps the representation of user and ad to each manifold. 
The origin of each manifold is set as the centroid of each user cluster.
The user preference for each ad is estimated using the distance between two entities in the hyperbolic space, and the final prediction is determined by aggregating the values calculated from the learned multiple manifolds. We evaluate our method on public benchmark datasets and a large-scale commercial messenger system LINE, and demonstrate its effectiveness through improved performance.

\end{abstract}


\maketitle

\section{Introduction}
\begin{figure}[!t]
  \centering
  \includegraphics[width=.7\linewidth]{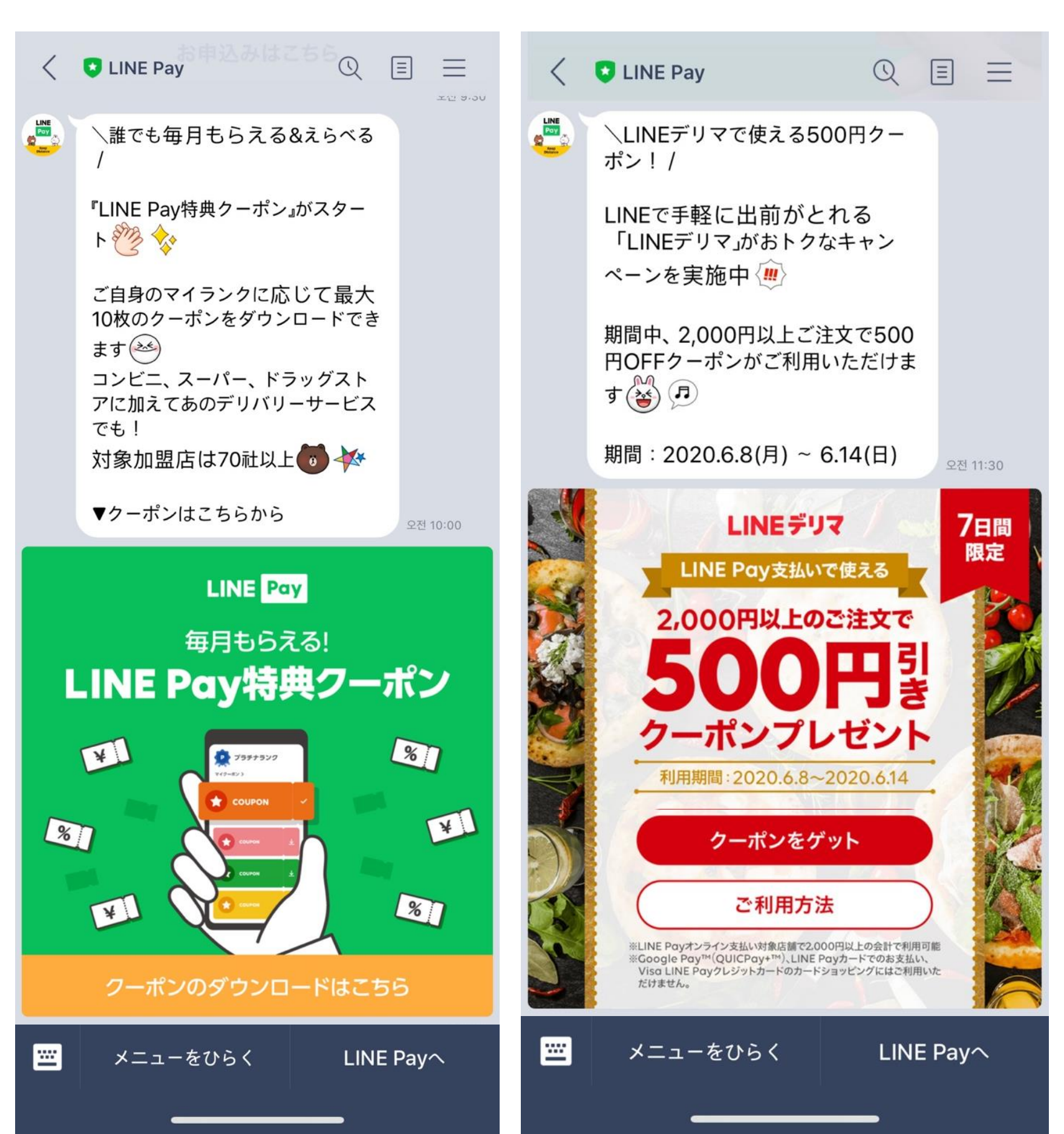}
  \caption{LINE messenger advertisement system.}
  \label{fig:Line}
\end{figure}

Messenger platform is an emerging advertisement channel.
In messenger platform, users experience a message-typed advertisement (ad) with a separate chat room feeling more private and direct compared to traditional ad channels, e.g., search engine, and web portal. 
High penetration ratios of smartphone and SNS utilization enable messenger ad system to become more promising with high sales~\citep{zhang2018mobile}.
\figurename~\ref{fig:Line} shows an example of our LINE messenger advertisement system.

However, an ad for broad random users without precise user targeting can not resonate with their potential audience playing as annoying spam. In this paper, since the accurate representation of the users and ads is a necessary for targeted advertising system, we present the deep representation learning scheme using hyperbolic geometries. Our method enables effective capture of the hierarchical and complex relationships between users and ads.

\begin{figure}[!t]
  \centering
  \includegraphics[width=.50\linewidth]{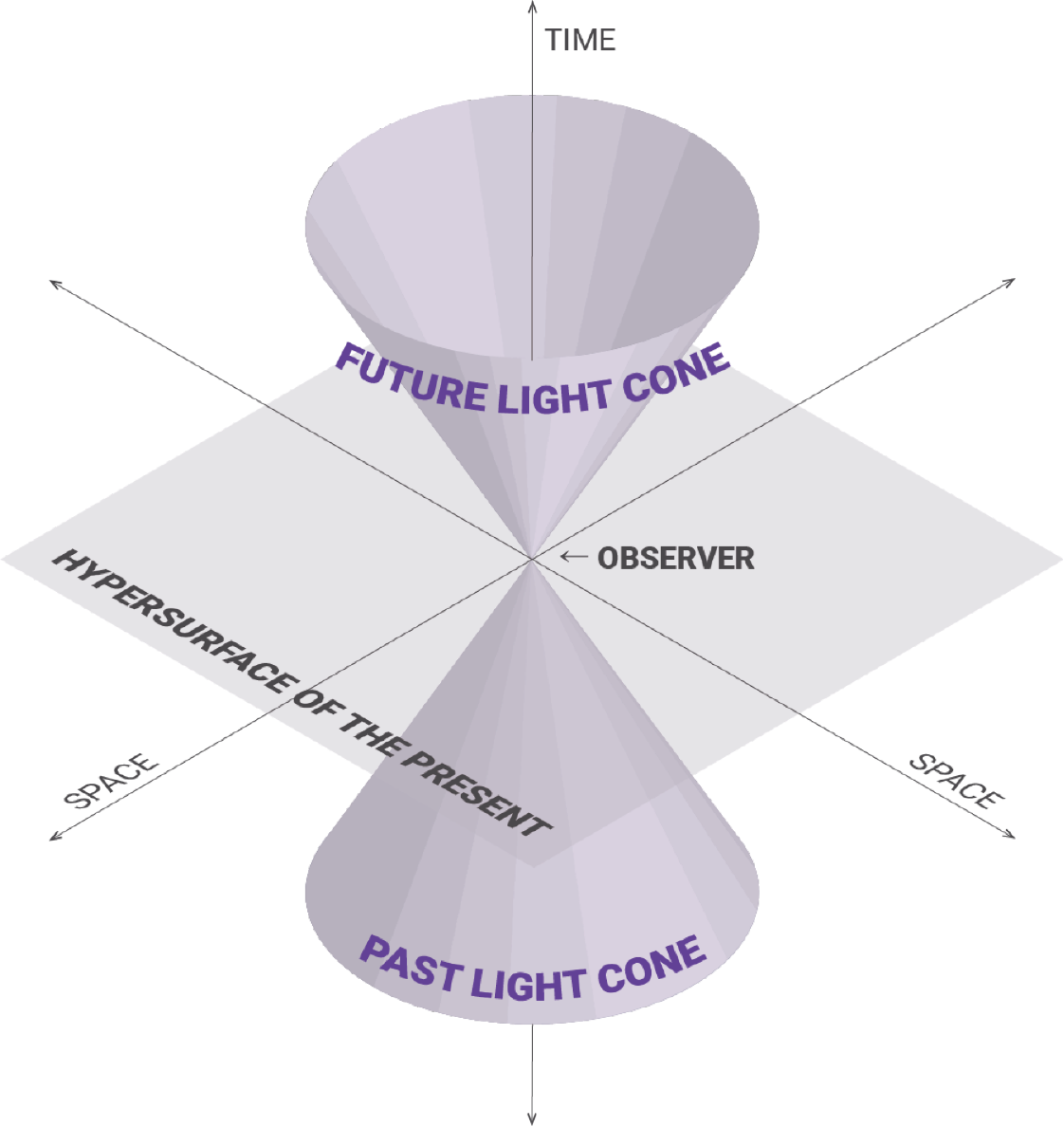}
  \caption{Minkowski space.}
  \label{fig:Minkowski}
\end{figure}

One of the most prominent approaches in traditional studies is Collaborative Filtering (CF)~\citep{schafer2007collaborative, zhang2016collaborative, chen2019collaborative} which finds a group of users who have responded to the ads similar to the target ad. 
Owing to the limitation of the low-dimensional representation of CF, recent studies have presented various neural network based approaches~\citep{kim2019tripartite, yi2019deep, zhang2019deep, park2020hop} that can effectively embed user and advertisements into the high-dimensional spaces. Despite the wide success and expansion of those methodologies, most of them implicitly embed the entities (i.e., users and ads) into points in Euclidean space, which causes an inherent limitation in the representation power. The latest studies, however, pointed out that the real-world user-item interaction datasets exhibits the hierarchical structures; therefore, it is more desirable to map embeddings into a hyperbolic space than Euclidean space~\citep{wang2015exploring, tran2018hyperml, schmeier2019music, chamberlain2019scalable}.
Unlike in the flat (Euclidean) plane, the distance between the nodes in tree-structured data is preserved in hyperboloid~\citep{gromov1987hyperbolic}, and therefore hyperbolic geometry is proven to naturally suitable for modeling hierarchical structures.

Although the hyperbolic space has successfully reflected the topology in user-item representation, the existing approaches fix its origin and use single manifold as an embedding space.
In a real-world, large-scale advertising system, there exist various groups of users with different preference characteristics, and it may not be valid to assume that every user and advertisement entity can be expressed by using single geometry.
There were several researches in Euclidean space that improves the prediction performance by adopting clustering algorithms into the recommendation tasks~\citep{ungar1998clustering, dubois2009improving, gong2010collaborative}.
However, usage of clustering scheme into the hyperbolic manifold learning on advertising system had not yet been reported.
The main contribution of this paper is to extend the hyperbolic representation learning to the \emph{Multi-Manifold Learning} framework by constructing multiple hyperbolic spaces centered on each clustered user group.

We evaluate the proposed framework on a large-scale real-world dataset collected from LINE messenger platform. The experimental results demonstrate that the proposed model increases the prediction performance and allows the representation to be diversified. We further report the performance on the public benchmark datasets to show that Multi-Manifold Learning can be applied commonly to various tasks as well as the targeted advertising.

\section{Targeted advertising system}
\label{sec:Recommender}

Unlike the traditional forms of advertising that expose ads to random users, the core of targeted advertising is that the system sends ads to different user groups based on the user-ad-preferences. The targeted advertising system uses the side information of those ads, such as images and advertising phrases, as well as the user's demographic information and click history to find the most relevant user group. 

Starting from the $X_{u}$ for users and $X_{a}$ for advertisements with attribute matrices $X_{u} \in \mathbb{R}^{N_{u} \times F_{u}}$ and $X_{a} \in \mathbb{R}^{N_{a} \times F_{a}}$, the neural networks $f_{u}: \mathbb{R}^{F_{u}} \rightarrow \mathbb{R}^{H}$ and $f_{a}: \mathbb{R}^{F_{a}} \rightarrow \mathbb{R}^{H}$ transforms the $X_{u}$ and $X_{a}$ to $Z_{u} \in \mathbb{R}^{N_{u} \times H}$ and $Z_{a} \in \mathbb{R}^{N_{a} \times H}$, where 
$F_{v}$, $N_{v}$, and $H$ denotes the number of attributes of $v \in \{u, a\}$, and the number of hidden features, respectively.
To build a more powerful user representations, we introduce additional neural networks $f_{h}: \mathbb{R}^{F_{a}} \rightarrow \mathbb{R}^{H}$ that embeds users' click history matrix $C$ into $Z_{h} \in \mathbb{R}^{N_{u} \times H}$ where $c_{i, j}$ is one if there is positive interaction between the $i$-th user and $j$-th advertisement and is zero otherwise.

Finally, the preference scores, $P_{u, a}$, between the users and ads are computed through the distance or inner-product between embeddings of them:
\begin{equation} \label{eq:prudict}
P_{u, a} = \text{Decision}(dist(Z_{u} + Z_{h}, ~~ Z_{a})),
\end{equation}
where $dist(p, q)$ is the distance between two points $p$ and $q$ on the given manifold.
User preference scores are sorted for each ad, and the top $k$ users with the highest scores are selected as the targeted users for the ad.
In this paper, we used Fermi-Dirac decoder~\citep{krioukov2010hyperbolic, nickel2017poincare} for the decision function.

\begin{figure}[!t]
  \centering
  \includegraphics[width=.5\linewidth]{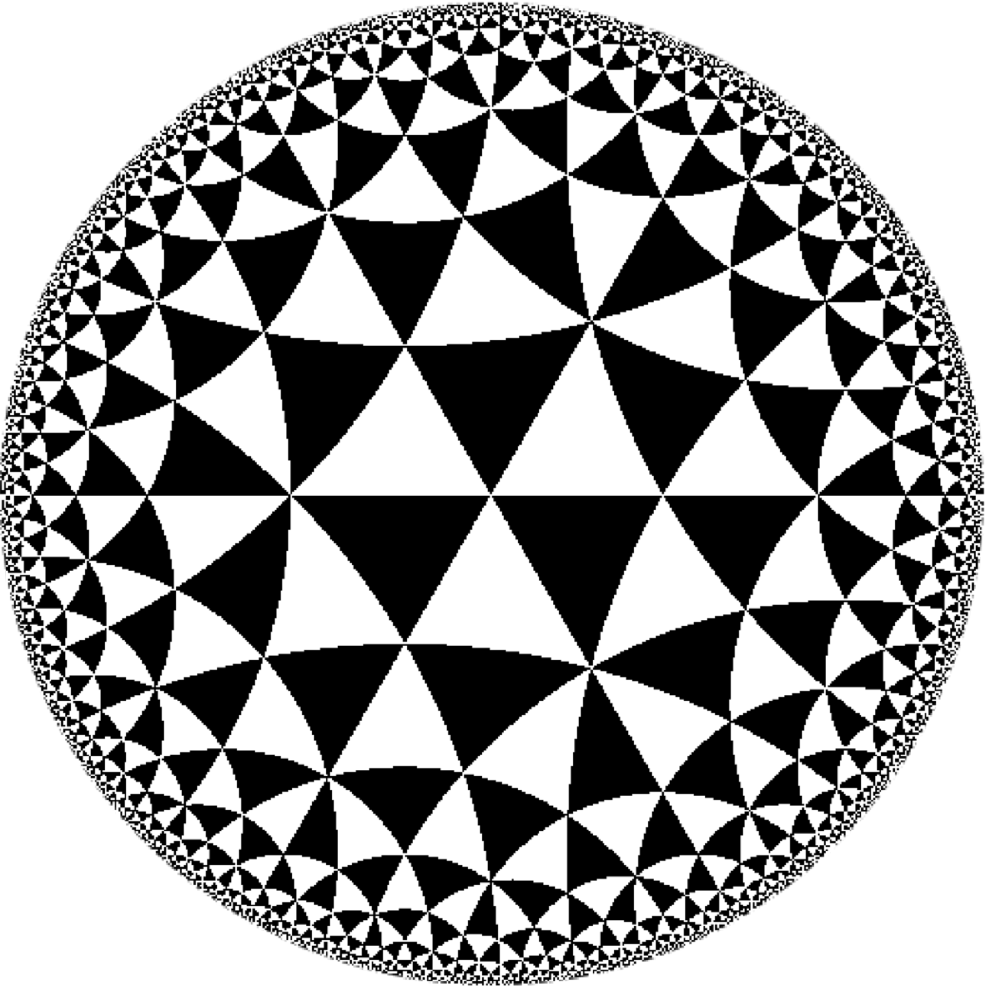}
  \caption{M.C. Escher style illustration of the \Poincare~disk model.}
  \label{fig:Escher}
\end{figure}

\section{Hyperbolic Geometry}
\subsection{Riemannian Manifolds}
\label{sec:Riemannian}
\begin{figure*}[!t]
  \centering
  \includegraphics[width=.9\linewidth]{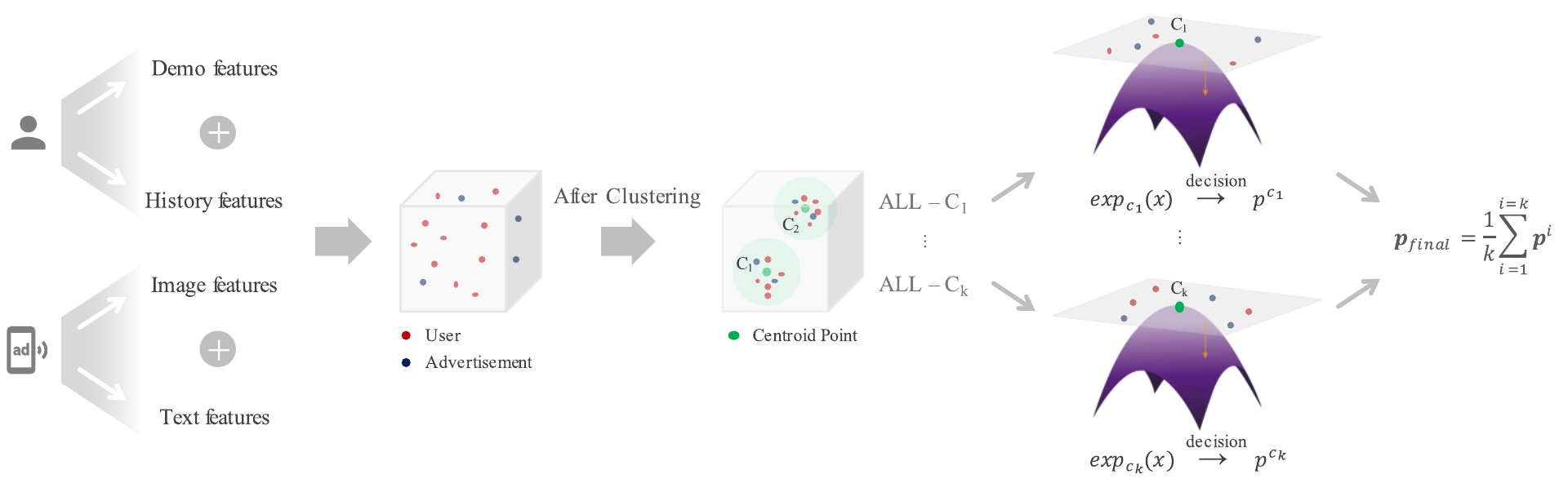}
  \caption{Conceptual scheme of our proposed method.}
  \label{fig:schematic}
\end{figure*}
A topological space $\mathcal{M}$ is a smooth manifold if $\mathcal{M}$ satisfies following four conditions: \textit{It is Hasudorff, It is second countable, $\mathcal{M}$ contained a open sets which is homeomorphic to $\mathbb{R}^{n}$, and its transition maps are infinitely differentiable}. For $p \in \mathcal{M}$, we can define the tangent space $T_p\mathcal{M}$ which is the first order approximation of $\mathcal{M}$ around point $p$. 

A Riemannian manifold is ($\mathcal{M}$, $g$), where $\mathcal{M}$ is a differential manifold and $g$ is Riemannian metric, which is a family of inner products on tangent spaces, $g_p : T_p\mathcal{M} \times T_p\mathcal{M} \rightarrow \mathbb{R}$ with smoothly varying $p$. Riemannian metric is used to measure distances by integrating the length between two points:
\begin{align} \label{distances}
d_g(p,q) = \text{inf}\int_{0}^{1} \sqrt{g_{\gamma(t)} (\dot{\gamma}(t), \dot{\gamma}(t))}
\end{align}
where $\gamma(0)=p$, $\gamma(1)=q$, and $\gamma \in C^{\infty}([0,1],\mathcal{M})$. A shortest path between two points $p$ and $q$ on curve $\gamma$ is called a geodesic, and equivalent to a straight line in Euclidean space.
From geodesic, we can define the projection by utilizing geodesic coordinates. This is called \textit{exponential map}  $exp_p$ at $p$, which projects a vector $v$ of the tangent space $T_p\mathcal{M}$ at $p$ to a point $exp_p(v) \in \mathcal{M}$ on the manifold. In this map, $\gamma$ is the unique geodesic satisfying $\gamma(0) \coloneqq p \in \mathcal{M}$ with unit-norm $\dot{\gamma}(0) \coloneqq v \in T_p\mathcal{M}$. Consequently, in very local area, exponential map is satisfying $exp_p(v) \coloneqq \gamma(1)$. The reverse map is called \textit{logarithmic map} that maps $q \in \mathcal{M}$ back to the tangent space $T_p\mathcal{M}$ at $p$ such that $log_p(exp_p(v)) = v$.

\subsection{Hyperbolic Space}
\label{sec:Hyperbolic}

Hyperbolic space is a non-Euclidean space with a constant negative Gaussian curvature. Gaussian curvature is the product of the principal curvature, which is divided into a sphere, hyperbola, and flat depending on whether the value is constantly positive, negative, or zero. Hyperbolic space is often associated with Minkowski spacetime in special relativity. Minkowski model is a $n$-dimensional hyperbolic geometry in which points are represented on the future light cone of a two-sheeted hyperboloid of $(n+1)$-dimensional Minkowski space as shown in~\figurename~\ref{fig:Minkowski}. 


\textbf{Learning on hyperbolic manifold.} Let $<\cdot,\cdot>_{g_M}:\mathbb{R}^{n+1} \times  \mathbb{R}^{n+1} \rightarrow \mathbb{R}$ denote the Minkowski inner product, $<p,q>_{g_M} \coloneqq -p_{0}q_{0}+p_{1}q_{1} \dots + p_{d}q_{d}$ with the coordinates $p_{0}$ and $q_{0}$ representing time. We denote $H^K$ as the hyperbolic manifold with constant negative curvature -$1/K$ $(K>0)$, and $T_pH^K$, the tangent space centered at point $p$. As described in Section~\ref{sec:Hyperbolic}, mapping between tangent space and manifold is performed by exponential and logarithmic maps. There are already known expressions of the exponential and the logarithmic maps on hyperboloid manifolds, which allow us to map points on hyperboloid to tangent spaces and vice-versa: For $p \in H^{K}, v \in T_pH^{K}$ and $q \in H^{K}$ such that $v\neq0$ and $q\neq p$, the exponential and logarithmic maps of the hyperbolic model are given by:
\begin{align} \label{exponential}
exp_{p}^{K}(v) = \text{cosh}(\frac{\norm{v}_{g_{M}}}{\sqrt{K}})p+\sqrt{K}\text{sinh}(\frac{\norm{v}_{g_{M}}}{\sqrt{K}})\frac{v}{\norm{v}_{g_{M}}}
\end{align}
\begin{align} \label{logarithmic}
log_{p}^{K}(q) = d_{g}^{K}(p,q)\frac{q+\frac{1}{K}<p,q>_{g_M}p}{\norm{q+\frac{1}{K}<p,q>_{g_M}p}},
\end{align}
where $\norm{v}_{g_M}=\sqrt{<v,v>_{g_M}}$ denotes norm of $v \in T_pH^{K}$ and \begin{math}d_g^{K}(p,q) = \sqrt{K}\text{arcosh}(-<p,q>_{g_M}/K)\end{math} denotes geodesic distance between $p$ and $q$. Above expressions assume that \begin{math}\gamma_{p \rightarrow v}^K(t) = \text{cosh}(\frac{t}{\sqrt{K}})p+\sqrt{K}\text{sinh}(\frac{t}{\sqrt{K}})v\end{math}, when $t$ is small enough and tangent vector $v$ is unit-speed, i.e. $<v,v>_{g_M}=1$.

\textbf{Diffeomorphism.} The hyperbolic model tends to be more robust and stable than the~\Poincare~model, but the~\Poincare~model is easier to interpret and can visualize embeddings directly on the~\Poincare~disk. Fortunately,~\Poincare~disk is a stereographic projection of hyperboloid~\citep{forrester2009derivation} which means theses two models are homeomorphic and exists a diffeomorphism $\Psi(\cdot)$ mapping hyperbolic model onto the~\Poincare~model: 
\begin{align} \label{diffeo}
\Psi(x_{0},x_{1}...,x_{d}) = \frac{\sqrt{K}(x_{1},x_{2}...,x_{d})}{x_{0}+\sqrt{K}},
\end{align} 
we will utilize deffeomorphism $\Psi(\cdot)$ for visualizing embeddings of data in~\figurename~\ref{fig:poincare}.

\textbf{Why hyperbolic manifold for targeted advertising.} The hyperbolic manifold is often considered as well-suited space for hierarchical structure. Suppose the task that embed a tree into the metric space while preserving its structural properties. i.e., the number of nodes at $l$-th layer is $n^l$. As a result, Euclidean space cannot contain all the nodes in the tree, which leads to poor representation of the model. However, in the hyperbolic space, the length of a circle is given as $2\pi$sinh$r$ with the constant Gaussian curvature $K=-1$. Since sinh$r$ = $\frac{1}{2}(e^r - e^{-r})$, the circle length grows exponentially with $r$, enough to include all the nodes. 
This property is illustrated in~\figurename~\ref{fig:Escher}. Each triangle has constant area in hyperbolic space, but in Euclidean space, it rapidly shrinks at the boundary.
The latest studies, pointed out that the real-world user-ads interaction exhibits the hierarchical relationships~\citep{nickel2017poincare, chami2019hyperbolic}; thus, the properties of hyperbolic space have great potential to learn distinct representations in targeted advertising system.

\section{Multi-Manifold Learning}

The core functionality of our large-scale targeted advertising system is to capture the representational differences between various user groups and advertisements. To do this, we propose \textit{Multi-Manifold Learning} that builds multiple manifolds for user groups because it may not be valid to assume that every user entity can be expressed by using single geometry. The conceptual scheme of our \textit{Multi-Manifold Learning} is shown in~\figurename~\ref{fig:schematic}.

Our proposed method consists of three stages. First, input $X_{u}$ and $X_{a}$ pass through DNNs $f_{u}$ and $f_{a}$ separately and users' click history $C$ passes through transformer network~\citep{vaswani2017attention} $f_{h}$.
Second, calculate user embedding $Z_{user}$ by adding $Z_{u}$ and $Z_{h}$, and cluster them into $T$ groups by using $k$-means clustering~\citep{alsabti1997efficient}.
We denote $c_t$ as a centroid of $t$-th group for $t \in \{1, \cdots, T\}.$
Finally, these embedding vectors map onto each $t$-th hyperbolic manifold of which the origin is $c_t$.
Then, we calculate the preference score on each manifold using a Fermi-Dirac decoder~\citep{krioukov2010hyperbolic, nickel2017poincare}, and aggregate them.

The detailed process can be formulated as follow:
\begin{align} \label{eq5}
E_{u}^{t} = (Z_{u} + Z_{h}) - c_{t},\;\;\;\; E_{a}^{t} = Z_{a} - c_{t} 
\end{align}
\begin{align} \label{eq6}
exp^K_{\textbf{o}}(E) = \Big{(}\sqrt{K}cosh(\frac{\norm{E}_{Euc}}{\sqrt{K}}),\sqrt{K}sinh(\frac{\norm{E}_{Euc}}{\sqrt{K}})\frac{E}{\norm{E}_{Euc}}\Big{)},  
\end{align}
where $E^t$ denotes embedding vector centered by centroid $c_t$.
The $exp^K_{\textbf{o}}(E)$ represents Euclidean vector $E$ mapped onto hyperbolic manifold with respect to the origin $\textbf{o}$. It is essential to centering the $Z$ with respect to centroid $c_t$ of each user group. Optimizing often fails if manifold's origin is set to a point with a value other than the origin $\textbf{o}$. The embeddings on each $t$-th hyperbolic manifold are used for computing user preference score through Fermi-Dirac decoder.
Finally, our overall probability and loss are:
\begin{align} \label{eq7}
p^{t}_{u,a} = \big{(}1+\exp^{(d_{g}^{K}(E^{t}_{u,~~h},  E^{t}_{a,~~h})-s)/b}\big{)^{-1}},\;\;\;\; E_{h} = exp^K_{\textbf{o}}(E)
\end{align}
\begin{align} \label{eq8}
\mathcal{L} = \sum \text{BCE}(P_{u,a}, Y),
\end{align}
where probability between user and ads on each manifold is $p^{t}_{u,a}$ and user preference of whole manifolds are  \begin{math}P_{u,a}=\frac{1}{t}\sum_{i=1}^{i=t}p^{i}_{u,a}\end{math}. The $s$ and $b$ in Fermi-Dirac decoder are hyper-parameter.

After mapping embeddings on the hyperboloid, an additional neural network layer such as Hyperbolic Neural Network (HNN)~\citep{ganea2018hyperbolic} can be added to perform weight learning on the hyperbolic manifold, but we empirically found that it does not show any performance improvements.

\setlength{\tabcolsep}{4pt}
\ctable[
    caption = {Model Performance on LINE messenger advertisement system},
    label = tab:performance,
     doinside = \footnotesize
]{lcccc}{
}{
\toprule
 & RocAuc   & Accuracy   & Average Precision & Shannon Entropy  \\
\midrule
CF  & 0.756 & 0.673  & 0.786 & 14.431  \\
MLP  & 0.770 & 0.681  & 0.775 & 14.197    \\
HNN  & 0.778 & 0.753  & 0.841 & 14.451    \\
\midrule
Multi-Manifold  & \textbf{0.818} & \textbf{0.765} & \textbf{0.846} & \textbf{14.567}  \\
\bottomrule
}

\setlength{\tabcolsep}{4pt}
\ctable[
    caption = {Model Performance on public benchmark MovieLens dataset},
    label = tab:performanceML,
     doinside = \footnotesize
]{lcccc}{
}{
\toprule
 & \multicolumn{2}{c}{MovieLens - 1M} & \multicolumn{2}{c}{MovieLens - 100K} \\
\cmidrule(l){2-3}  \cmidrule(l){4-5}
 & RocAuc   & Average Precision   & RocAuc  & Average Precision \\
\midrule
CF & 60.3 & 67.4  & 60.5 & 61.1  \\
MLP  & 57.4 & 66.3  & 61.7 & 62.0  \\
HNN   & \textbf{61.7} & 69.0  & 68.0 & 67.8    \\
\midrule
Multi-Manifold  & 61.5 & \textbf{69.8} & \textbf{68.3} & \textbf{68.5} \\
\bottomrule
}

\section{Experiment}
\subsection{Dataset}
\label{sec:Dataset}
\setlength{\tabcolsep}{5pt}
\ctable[
    caption = {Model performance comparison as the number of cluster increases on LINE messenger dataset.},
    label = tab:ablation,
     doinside = \footnotesize
]{rcccc}{
}{
\toprule
\# of Clusters & RocAuc   & Accuracy   & Average Precision & Shannon Entropy  \\
\midrule
1-cluster & 0.798 & 0.715 & 0.817 & 13.871   \\
3-cluster & 0.805 & 0.753 & 0.840 & 14.146   \\
5-cluster & \textbf{0.818} & \textbf{0.765} & \textbf{0.846} & 14.567    \\
10-cluster & 0.813 & 0.753 & 0.841 & 14.653 \\
15-cluster & 0.810 & 0.753 & 0.841 & \textbf{14.665} \\
\bottomrule
}

\begin{figure*}
\centering
\subfigure[MLP model]{\label{fig:a}\includegraphics[width=0.24\textwidth]{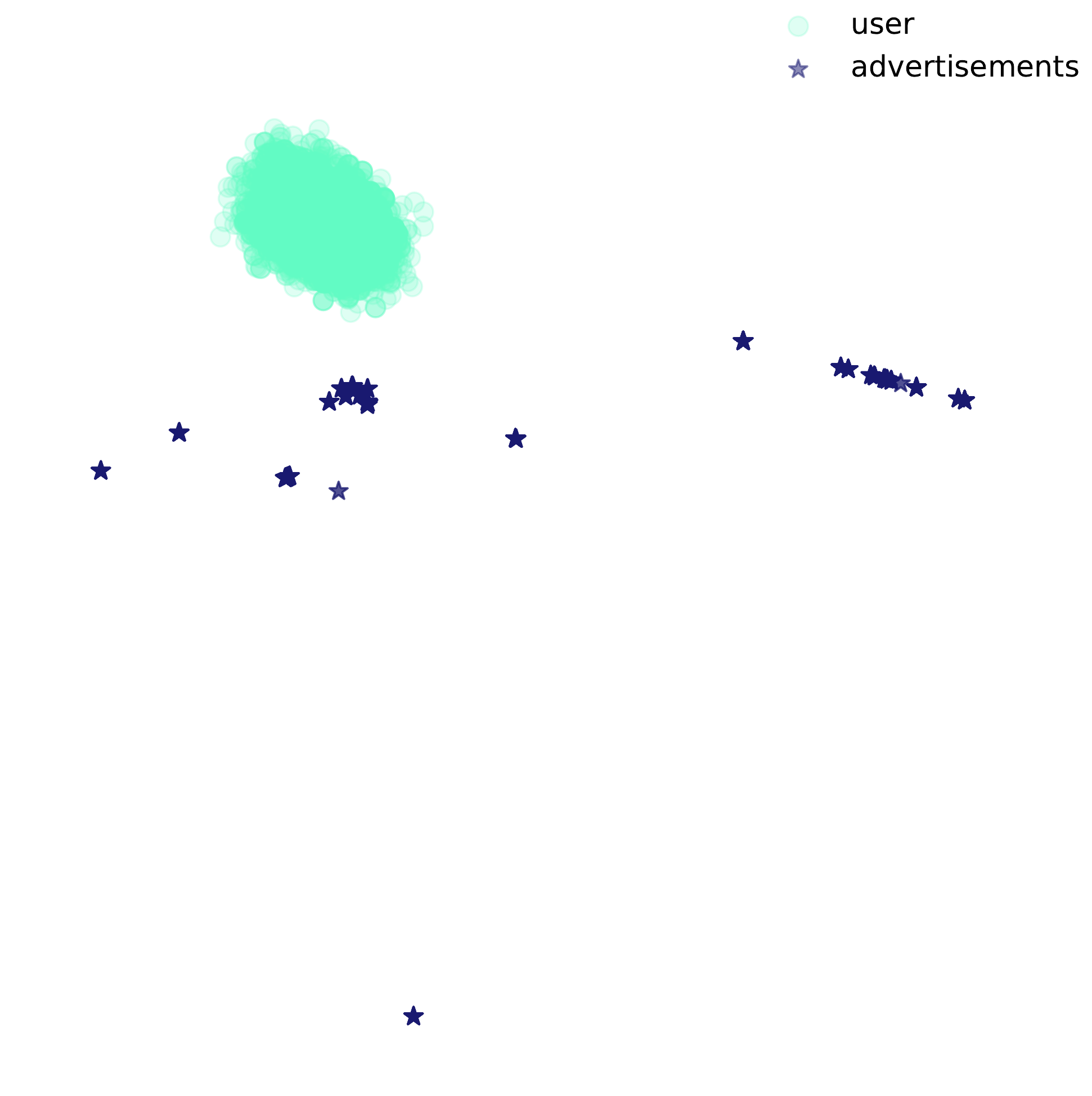}}
\subfigure[HNN model]{\label{fig:b}\includegraphics[width=0.24\textwidth]{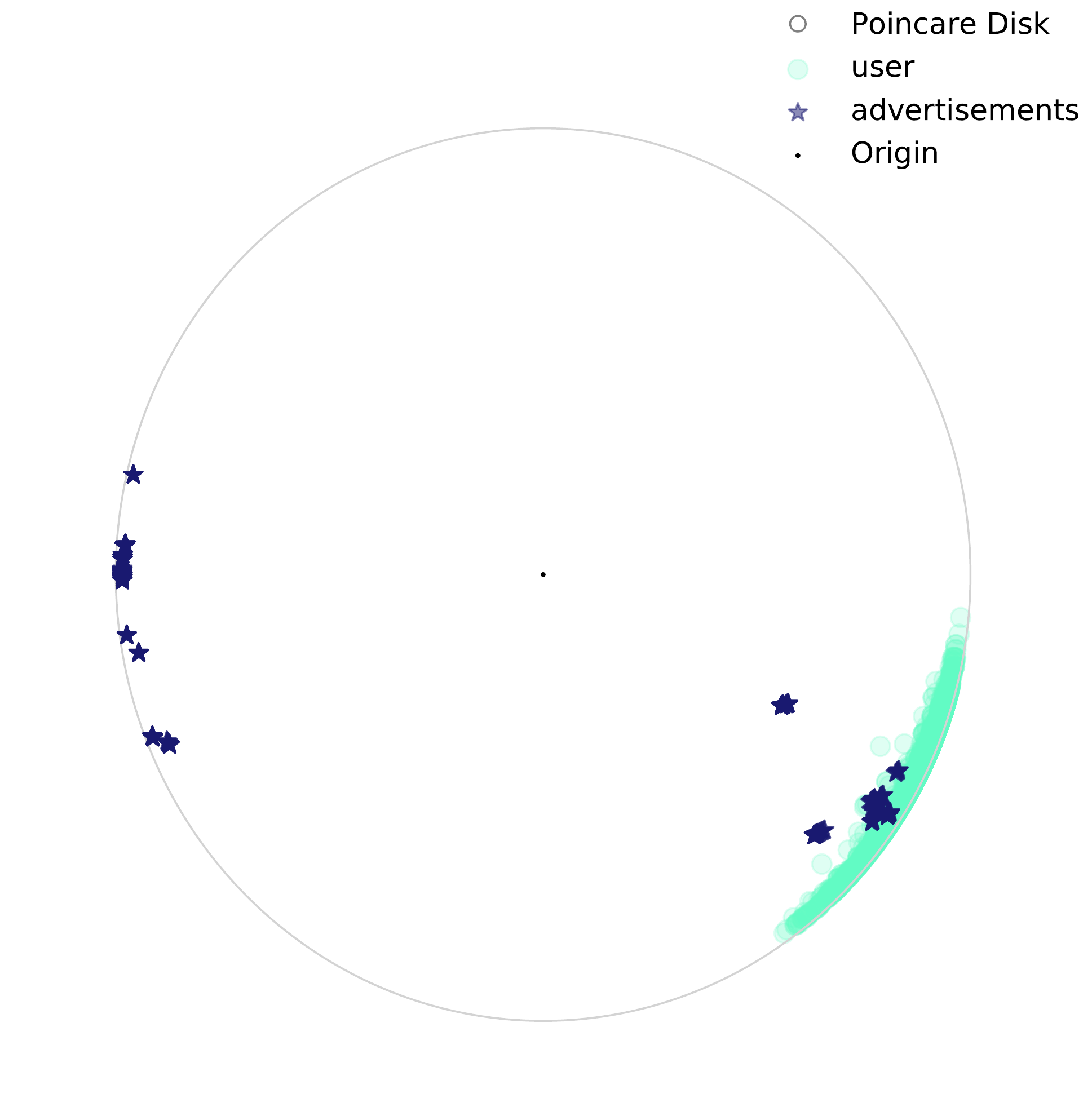}}
\subfigure[Our Multi-Manifold Learning]{\label{fig:c}\includegraphics[width=0.48\textwidth]{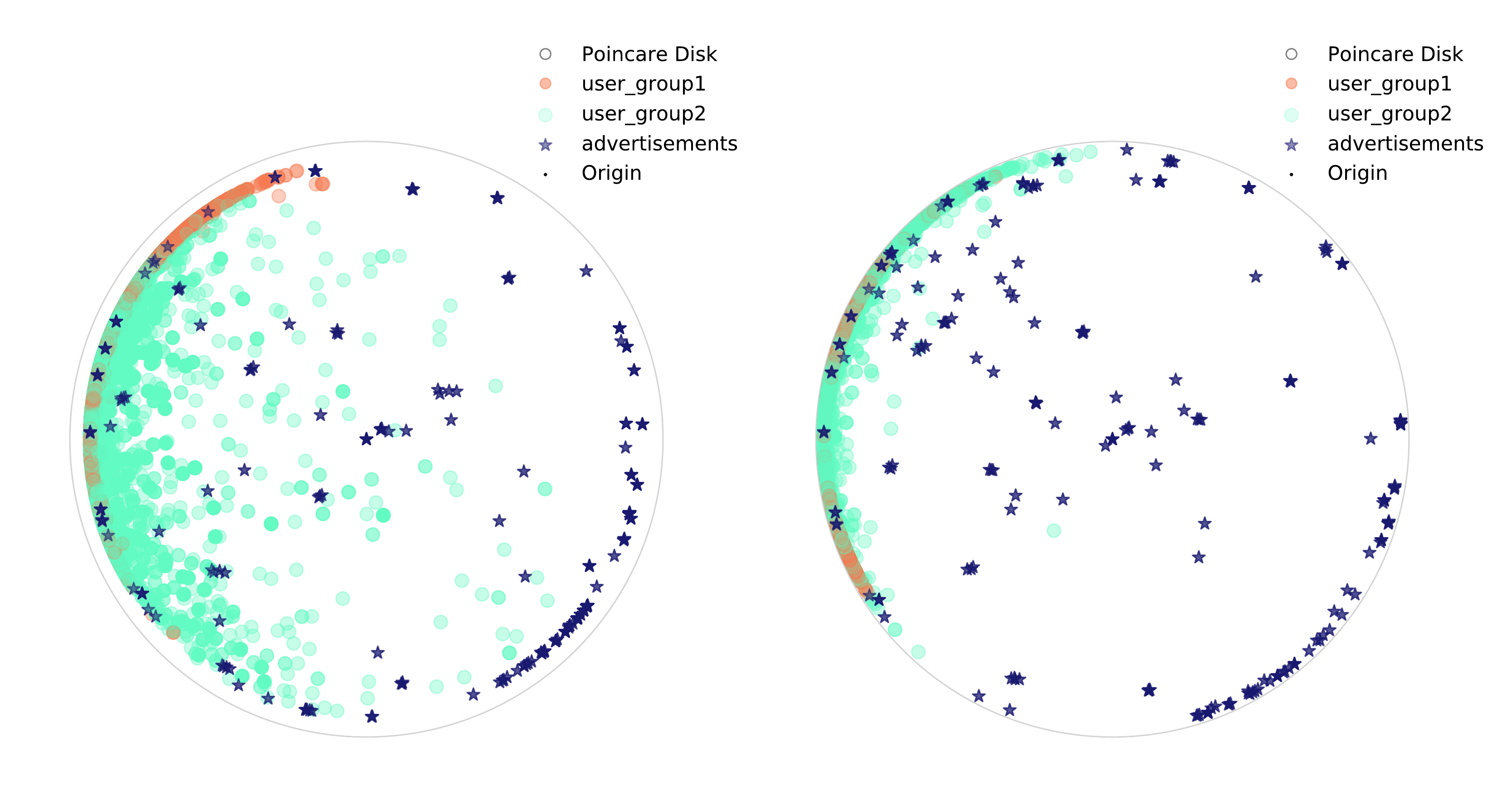}}
\caption{Visualization in embedding representations of users and advertisements. \textbf{(a)} embedding of Euclidean MLP, \textbf{(b)} \Poincare~disk visualization of HNN, \textbf{(c)} \Poincare~disk visualization of our Multi-Manifold Learning with two clusters on each manifold. We visualize them by using diffeomorphism between Hyperbolic space and~\Poincare~space as described in Section~\ref{sec:Hyperbolic}.}
\label{fig:poincare}
\end{figure*}
We collect dataset from LINE messenger platform that targets users from all over the world, and the number of users in the service is about 200 million. We randomly select one million users\footnote{Since the number of users using the service is huge, we use a subset of users for experiments.}
We split a dataset based on time: the first fourteen days for training and the subsequent two days for test. We report the performance of the last epoch.

For better representation of user and advertisement embeddings, age, gender, mobile OS type, interest, number of LINE Pay membership follower, and number of LINE Pay membership followee attributes are used for users, while text and image are used for advertisements. For each attribute, we use shallow DNNs to make $H-$dimensional feature vectors and aggregate them to get the feature matrices $Z_{u}$ and $Z_{a}$.

Due to a large number of users, we use 10,240 randomly sampled users for each batch. The advertisement click history is used up to the day before the forecast date, and click histories are normalized for each user. 

For a fair comparison with the base models, we extend our experiments to public benchmark datasets: MovieLens\footnote{https://grouplens.org/datasets/movielens}, which is widely used public dataset for recommender systems. We modify the dataset to binary classification: a label as 1 if the movie score is greater than 4, otherwise as 0.

\subsection{Baselines}
\label{sec:Comparable}
To demonstrate the effectiveness of proposed model, we compared our model with following three base models:

\begin{itemize}[leftmargin=*]
\item~\textbf{Collaborative Filtering (CF)}~\citep{schafer2007collaborative, zhang2016collaborative, chen2019collaborative}: The underlying assumption of Collaborative Filtering is the premise that users' past trends will remain the same in the future. In other words, it is a technique to identify users with similar patterns based on their preferences and interests.
\item~\textbf{Multilayer Perceptron (MLP)}~\citep{xue2017deep, yi2019deep, zhang2019deep}: 
There are numerous types of MLP algorithms that are based on Matrix Factorization.
We report the presented framework using Euclidean space as MLP in the following results.
Note that, Multi-Manifold Learning in Euclidean space is not reported, since the relative distance between two points is translation-invaraint in Euclidean space.
\item~\textbf{Hyperbolic Neural Network (HNN)}~\citep{ganea2018hyperbolic}: 
This work generalizes the linear transform and bias addition of DNNs on the hyperbolic space and proposes several important deep learning tools on the hyperbolic space. We use HNN, which is based on basic DNNs, where the core operations are executed in hyperbolic space. 
\end{itemize}

For the fairness of the comparison, we adopt the same neural network architectures for $f_{u}$ and $f_{a}$.
The hidden vector size is set to 64 and we do not use dropout~\citep{srivastava2014dropout} and l2 regularization. All the experiments were performed on NAVER SMART Machine Learning platform (NSML)~\citep{sung2017nsml, kim2018nsml} using PyTorch~\citep{NEURIPS2019_9015}.

\subsection{Experimental Results and Analysis}
\label{sec:Experimental}

We report three accuracy metrics of RocAuc, Accuracy, and Average Precision, and one diversity metric of Shannon Entropy. In particular, Average Precision is set up for the targeted Advertising System. The user preference is sorted for a specific advertisement, and then precision is calculated for each ad of the top $k$ users. Finally, we average the precision of all ads.

\textbf{Performance comparison.} As shown in Table~\ref{tab:performance}, our method shows the best prediction performance for all accuracy metrics, as well as the diversity metric. The diversity metric of Shannon entropy for each model shows how diversified the recommended users are. Our model shows the highest diversity compared to other baselines, indicating that superior expressiveness of embedding enables precise targeting. To further demonstrate the effectiveness of our model on general dataset, we present additional experimental results on public benchmark dataset MovieLens. 
As represented in Table~\ref{tab:performanceML}, our model shows the best or second-best performance, demonstrating its generality not overfitted to a certain dataset.

\textbf{Effects of the number of clusters.}
To illustrate the effect of the number of clusters on the model performance, we report the prediction accuracy and diversity of our model for different $T$'s.
Table~\ref{tab:ablation} shows that the overall performance improves as the cluster grows, and the best performance is obtained at $T=5$.
After $T=5$, the prediction accuracy converges while the diversity improves further.
Overall, we select the $T$ as five throughout the experiments.

\textbf{Embedding visualization.} \figurename~\ref{fig:poincare} shows how Multi Manifold Learning works compared to others. From embedding visualization of MLP model and HNN model, we can verify their positive user pool responding to ads are very small. On the other hand, our proposed method Multi-Manifold Learning shows our model includes many users in a positive pool that is compatible with ads. 

The data embedding in a different hyperboloid, originating from centroids of different user groups, have different embedding spaces. Since we set two clusters, the~\figurename~\ref{fig:poincare} shows results for two manifolds. We can verify each manifold has a different distance in the hyperbolic space, ads that are not relevant to the user are pushed to the edge, while preferred ads appear to move toward the center. It is because the hyperbolic space we constructed is centered by the centroids of the well-clustered user group.

\section{CONCLUSION}

Traditional targeted advertising systems struggle with data representation capabilities because of the inherent limitation of Euclidean space. To tackle this issue, we present Multi-Manifold Learning, a well-designed technique to learn better representation of users and advertisements. Experimental results show the proposed scheme improves the targeted advertising quality in terms of both accuracy and diversity. As the future directions, we will develop a Multi-Manifold Learning scheme in terms of diffeomorphism learning. Besides, we will extend our method on real-world large scale online service of LINE messenger platform.

\begin{acks}
The authors would like to thank Professor Hyunwoo J. Kim and NAVER Clova ML X team for insightful comments and discussion.
\end{acks}

\bibliographystyle{ACM-Reference-Format}
\bibliography{references}

\end{document}